\documentclass[preprint,12pt]{elsarticle}

\usepackage{amssymb}

\journal{Fuzzy Sets and Systems}

\begin{document}

\begin{frontmatter}



\title{Using Memristor-Crossbar Structure to Implement a Novel Adaptive Real-time Fuzzy Modeling Algorithm}


\author{Iman~Esmaili Paeen Afrakoti,\ Saeed~Bagheri Shouraki\\
and Farnood~Merrikhbayat}

\address{Research Group of Brain Simulation and Cognitive Science, Artificial Creatures Lab, Electrical Engineering School, Sharif University of Technology, Azadi Avenue, Tehran, 11365-9517, Iran}

\begin{abstract}
Although fuzzy techniques promise fast meanwhile accurate modeling and control abilities for complicated systems, different difficulties have been revealed in real situation implementations. Usually there is no escape of iterative optimization based on crisp domain algorithms. Recently memristor structures appeared promising to implement neural network structures and fuzzy algorithms. In this paper a novel adaptive real-time fuzzy modeling algorithm is proposed which uses active learning method concept to mimic recent understandings of right brain processing techniques. The developed method is based on processing fuzzy numbers to provide the ability of being sensitive to each training data point to expand the knowledge tree leading to plasticity while used defuzzification technique guaranties enough stability. An outstanding characteristic of the proposed algorithm is its consistency to memristor crossbar hardware processing concepts. An analog implementation of the proposed algorithm on memristor crossbars structure is also introduced in this paper. The effectiveness of the proposed algorithm in modeling and pattern recognition tasks is verified by means of computer simulations. 

\end{abstract}

\begin{keyword}

Fuzzy inference\sep Active Learning Method \sep Optimization-free \sep Memristor-crossbar \sep Pattern classification

\end{keyword}

\end{frontmatter}


\section{Introduction}
During more than five decades of the fuzzy control and modeling methods being introduced, many efforts have been conducted in search for effective methods to overcome the crisp domain difficulties in modeling and control applications. These difficulties arise mostly from fitting an exact mathematical relationship among system inputs, state variables and system outputs. Humans possess a remarkable ability to process intricate information with ease. fuzzy logic is modeled on the linguistic and logical aspects of the human thought processes so involves robust computing in the presence of uncertainty\cite{zadeh1,zadeh2}. However, due to the fact that achieving sufficient accuracy requires the use of iterative optimization techniques which are mainly based on crisp domain calculations, using them in fuzzy aspects cannot be avoided\cite{sugeno,tanaka}. 

During mid 90s, researchers tried incorporate hardware implementations of the human understanding based on physically models of the right-brain processing\cite{shibata,degaris}, in order to overcome the aforementioned difficulties. 
Realization of the memristor in 2008 \cite{strukov}, and the similarities between its functional behavior and the synaptic processing of the brain, opened a new path for the brain emulation researches. Initial reports on the matter, proposed some memristor-crossbar based implementations of the spiking neural network structures with effectiveness in clustering tasks \cite{bernabe1,bernabe2}. Later in 2011, a mixed analog and digital soft computing system known as active learning method (ALM) based on memristor crossbar structures, was proposed\cite{farnood1}. ALM is a fuzzy modeling technique that mimics some understanding of the information handling processes of the human brain. This stable and fast converging technique uses image information for modeling task, without including complex mathematical expressions\cite{alm1}. The proposed hardware was very huge with high complexity in the feature extraction phase. In 2012, the combination of the memristor crossbar with fuzzy logic to create an analog memristive neuro-fuzzy computing system with fuzzy input and output terminals based on Hebbian learning rule was proposed\cite{farnood2}. For achieving more accuracy in these two latter systems an optimization algorithm had to be applied. This makes the algorithms inefficient for the real-time applications. If the optimization algorithms could be eliminated, the fuzzy processing operations would be effectively separated from the crisp domain, even in the calculation phase.
 
In this paper a novel optimization-free fuzzy modeling algorithm with plasticity, fast convergence and stable characteristics is proposed that can be implemented directly on the memristor-crossbar structure. The model is constructed using the input-output data sample gathered from the system and the output will be computed using the max-min inference algorithm. The output fuzzy number can be used as an input of another fuzzy modeling block or be translated using a defuzzication algorithm to a crisp number for the real world applications. This algorithm has the advantage of simplicity, and it is optimization-free. A fully analog hardware for the implementation of the algorithm is proposed.

The rest of the paper is organized as follows. A brief overview of active learning method is presented in Section 2. In Section 3 the structure of the proposed algorithm is shown. The inference algorithm of the proposed method is introduced in section 4. Section 5 gives a brief introduction to memristive devices.
Hardware implementation of the proposed algorithm is discussed in section 6. Simulation results are provided and discussed in detail in Section 7. Finally conclusions and discussions are drawn in Section 8.

\section{Active Learning Method}
Active Learning Method (ALM) was inspired by the way humans behave when confronting a complex problem\cite{alm1}. The main idea of ALM is to approximate a Multiple Inputs-Single Output (MISO) system with several Single Input-Single Output (SISO) subsystems. Each subsystem shows the overall behavior of output with respect to one input; the behavior will be extracted from the pattern generated on a two dimensional plane (Image) using the Ink Drop Spread (IDS) operator. The IDS operator is applied to data points on corresponding plane with the ink stains diffusing and aggregating, resulting in a smooth pattern representing a single variable functional behavior. 

\begin{figure}
\centering
\includegraphics[scale=0.4]{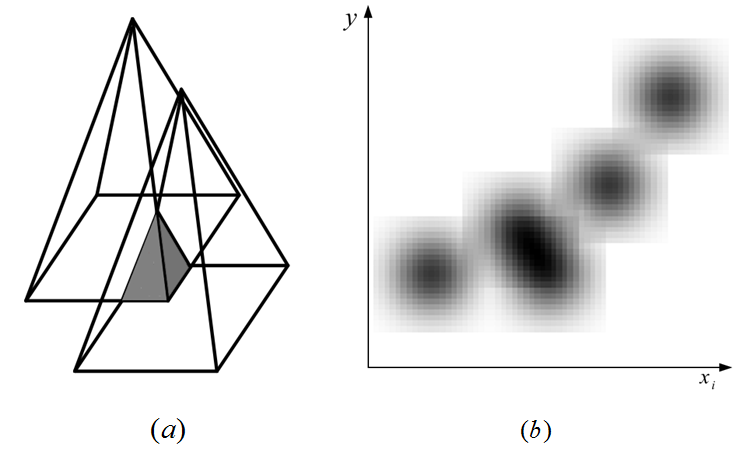}
\caption{a) Ink stains with pyramid shape; b)Five Inks are diffused on the plane.}
\label{ink}
\end{figure}

The pattern on a plane is constructed by projecting all $ (x_i,y) $ data on it. Diffusion of information has the main role in this process; common sense implies that each sample data not only has information in its exact point but also  is  valid in neighboring points with less confidence degree. By getting away from the point of sample data, the confidence degree will decrease. This is similar to the notion of fuzzy membership functions. The diffusion of information can thus be implemented by considering a three dimensional pyramid-shape fuzzy membership function centering at each data point as shown in figure \ref{ink}. Figure \ref{ink}(a) depicts two such membership functions overlapped and aggregated. An IDS plane after applying IDS operator to the five data samples is shown in figure \ref{ink}(b). For the sake of simulation and hardware conformity, the range of input variables should be quantized to $ n $ levels, giving rise to $ n\times n $ IDS planes.

Although ALM has excellent performance in many applications like function modeling \cite{almmod}, classification \cite{almclas} and control \cite{almcon1,almcon2} , it needs to divide the input variables range into many intervals; finding the suitable place of division is always a big concern and needs an time consuming optimization algorithm \cite{almop1,almop2}.

\section{proposed algorithm}
For understanding the structure of the proposed algorithm we are going to introduce two concepts. The first concept is IDS plane. Each IDS plane is a plane whose  horizontal axis is one of the input variables and its vertical axis is the output variable the  same as the one in ALM. Each training data, $ (x_i,y) $ should be diffused on an IDS plane.

\begin{figure}
\centering
\includegraphics[scale=.5]{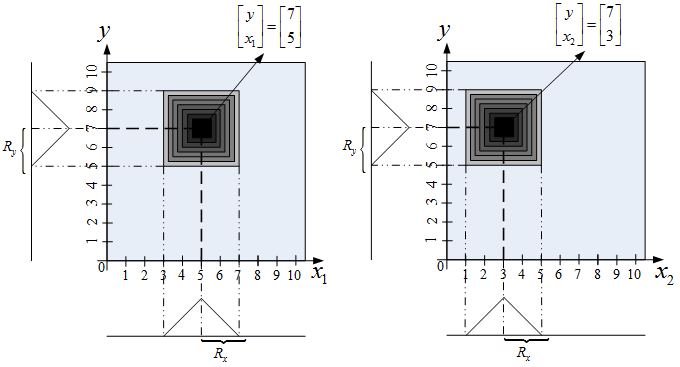}
\caption{An instant IDS group for $(x_1,x_2,y)=(5,3,7)$.}
\label{ID group}
\end{figure}

The next concept is IDS group. For a system with $ n $ inputs variable and one output, $ n $ IDS planes is needed for diffusing the data $ (x_1,y), (x_2,y),\ . . .\ , (x_n,y) $ on them using IDS operator. In this paper we named each group of $ n $ IDS planes as a IDS group. In figure \ref{ID group}, the sample data $(x_1,x_2,y)=(5,3,7)$, for a system with two inputs and one output, is diffused on an IDS group, as shown in \ref{ID group} two IDS planes are used for constitution of the related IDS group and the inks radius is two in each direction.

\begin{figure}
\centering
\includegraphics[scale=.7]{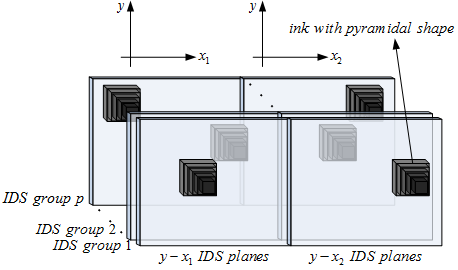}
\caption{The constructed IDS groups for a two inputs and one output system with $ p $ training data.}
\label{ID group1}
\end{figure}

After introducing these two concepts, the structure of the algorithm can be introduced. Assume that there is $ p $ sample data for a system with $ n $ inputs and one output. One IDS group is defined for each sample data, so each sample data will be diffused on the IDS planes of one IDS group using the IDS operator. As a result there will be $ p $ IDS group that simulate the system. In figure \ref{ID group1} the structure of the IDS groups is shown for a system with two inputs and one output with $ p $ training data; each IDS group  consists of two IDS planes that the pair $ (x_i, y), i=1,2 $ data is diffused on the related IDS plane.

\section{Inference algorithm of the proposed structure}
\begin{figure}
\centering
\includegraphics[scale=.5]{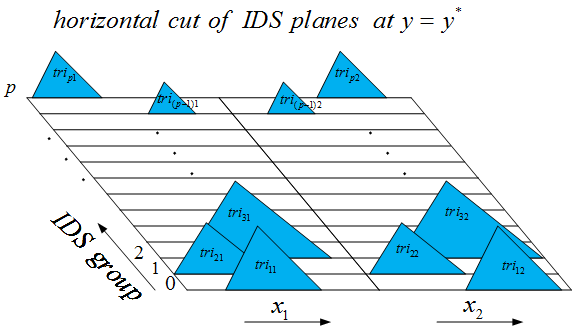}
\caption{The plane which is generated using horizontal cut of the IDS groups.}
\label{horizontal_page}
\end{figure}

For deriving the inference algorithm we are going to analyze a plane which is constructed with a horizontal cut of the IDS groups in $ y=y^* $  named $ y^* $ plane. In figure \ref{horizontal_page} a sample horizontal cut plane is shown. Each triangle, $ tri_{ij} $, shows the amount of darkness on $\textit{j}$th IDS plane of the $\textit{i}$th IDS group at $ y=y^* $. Each point on any triangle is a confidence degree to the $ y^* $ value of the output variable when the relative input is in the triangle range. Now we are going to write a rule of inference from this horizontal cut of IDS groups. Assuming a row of this plane, the triangles on any row are  inserted if and only if the sample data $ (x_{11}^*, x_{12}^*, y^*) $ or its neighboring data have occurred. This can be translated to the following fuzzy rule: $ if \ x_1 \ is \ tri_{11} \ and \ x_2 \ is \ tri_{12} \ then \ y=y^* $. This is a rule just for one row of this plane. Other rows will have the same rule which will be executed parallel to this rule; speaking in fuzzy literature, these rules have OR relation to each other. So the fuzzy $ if \ then \ rule $ for this cut can be written as Eq. (\ref{2}).
 
$ R_{y^*}: $
\begin{eqnarray}
\label{2}
& if & \ \ x_1 \ \ is \ \ tri_{y^*11} \ \ AND \ \ x_2 \ \ is \ \ tri_{y^*12} \ \ OR \nonumber \\
& x_1 & \ \ is \ \ tri_{y^*21} \ \ AND \ \ x_2 \ \ is \ \ tri_{y^*22} \ \ OR \ \ ... \nonumber \\
& x_1 & \ \ is \ \ tri_{y^*p1} \ \ AND \ \ x_2 \ \ is \ \ tri_{y^*p2} \ \ then \ \ y=y^* \nonumber \\
\end{eqnarray}

Where in this rule $ tri_{y^*ij} $ is the triangle $ tri_{ij} $ on the $ y^* $ plane. If the resolution of the output variable, $ y $, is set to $ n_y $, (i.e. it is quantized into $ n $ level) then there will be $ n_y $ rule like $ R_{y^*} $. In Eq. \ref{3} the complete rule base of a system with two inputs and one output and $ p $ training data is shown.\\
$ R_1: $
\begin{eqnarray*}
& if & \ \ x_1 \ \ is \ \ tri_{y_111} \ \ AND \ \ x_2 \ \ is \ \ tri_{y_112} \ \ OR \\
& x_1 & \ \ is \ \ tri_{y_121} \ \ AND \ \ x_2 \ \ is \ \ tri_{y_122} \ \ OR \ \ ...\\
& x_1 & \ \ is \ \ tri_{y_1p1} \ \ AND \ \ x_2 \ \ is \ \ tri_{y_1p2} \ \ then \ \ y=y_1
\end{eqnarray*}
$ . $\\
$ . $\\
$ . $\\
$ R_t: $
\begin{eqnarray*}
& if &\ \ x_1 \ \ is \ \ tri_{y_t11} \ \ AND \ \ x_2 \ \ is \ \ tri_{y_t12} \ \ OR \\
&x_1 &\ \ is \ \ tri_{y_t21} \ \ AND \ \ x_2 \ \ is \ \ tri_{y_t22} \ \ OR \ \ ...\\
&x_1 &\ \ is \ \ tri_{y_tp1} \ \ AND \ \ x_2 \ \ is \ \ tri_{y_tp2} \ \ then \ \ y=y_t \\
\end{eqnarray*}
$ . $\\
$ . $\\
$ . $\\
$ R_{n_y}: $
\begin{eqnarray}
\label{3}
&if& \ \ x_1 \ \ is \ \ tri_{y_{n_y}11} \ \ AND \ \ x_2 \ \ is \ \ tri_{y_{n_y}12} \ \ OR \nonumber \\
&x_1& \ \ is \ \ tri_{y_{n_y}21} \ \ AND \ \ x_2 \ \ is \ \ tri_{y_{n_y}22} \ \ OR \ \ ...\nonumber \\
&x_1& \ \ is \ \ tri_{y_{n_y}p1} \ \ AND \ \ x_2 \ \ is \ \ tri_{y_{n_y}p2} \ \ then \ \ y=y_{n_y} \nonumber \\
\end{eqnarray}

By introducing the $ if \ then $ rules of the system, the inference algorithm can be easily introduced. The confidence degree in the antecedent part of each rule is computed using both S-norm and T-norm operators. In this paper the minimum and maximum operators are chosen for T-norm and S-norm respectively. The output of each rule will be a pair, $ (y_i,\mu(y_i)) $ , which $ y_i $ is a quantized level of output variable and $ \mu(y_i) $ is the system confidence degree to this quantized level. So for a system with $ n_y $ quantized level there will be $ n_y $ pairs like this one. By putting together these pairs we get the final fuzzy output which should be transformed to a crisp number using a suitable defuzzifier algorithm for applying to the real world applications. In this paper we used the weighted sum formula (WSF) defuzzifier as shown in Eq. (\ref{4}).
\begin{equation}
\label{4}
y_{out}=\frac{\sum \limits_{i=1}^n y_i \times \mu_{y_i}}{\sum \limits_{i=1}^n \mu_{y_i}}
\end{equation}

\begin{figure}
\centering
\includegraphics[scale=.37]{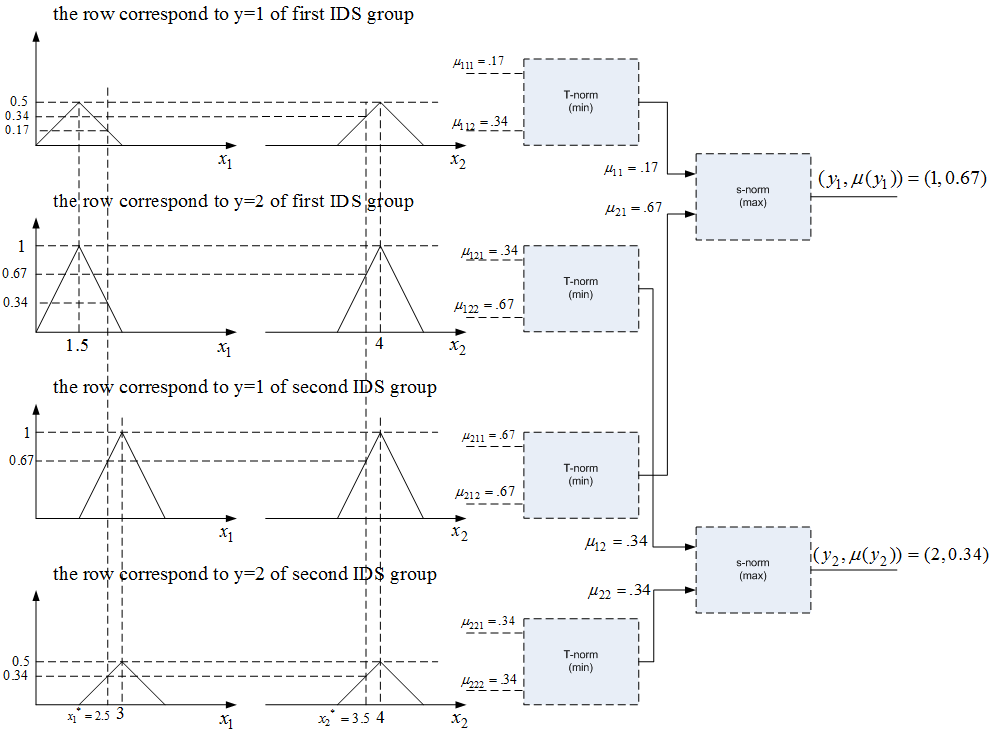}
\caption{The whole inference procedure for a simple problem.}
\label{procedure}
\end{figure}

In order to show the overall procedure of algorithm, a simple example is shown in figure \ref{procedure}. Here a system with two input and one output variables is assumed. For two training sample data there will be two IDS groups each with two IDS planes. And for simplicity it is assumed that the resolution of the output variable is set to 2, so there will be two rows for each IDS plane. It is assumed that the training data are  $ (x_1,x_2,y)=(1.5,4,2) $ and $ (x_1,x_2,y)=(3,4,1) $. The radius of ink stains for the output variable is set to 1.5. Now assume that the output should be computed  for the input data $ (x_1,x_2)=(2.5,3.5) $  using the proposed inference algorithm. In figure \ref{procedure} $ \mu_{kij} $ is the confidence degree of the $ \textit{t} $h IDS plane from the $ \textit{k} $th IDS group to the $ \textit{i} $th quantized level of the output variable, $ \mu_{st} $ is the confidence of the $ \textit{s} $th IDS group to the $ \textit{t} $th quantized level of the output variable. As can be seen in figure \ref{procedure} the confidence degree of IDS planes to the quantized levels of output variable are $ \ \mu_{111}=0.17,\ \mu_{112}=0.34,\ \mu_{121}=0.34,\ \mu_{122}=0.67,\ \mu_{211}=0.67,\ \mu_{212}=0.67,\ \mu_{221}=0.34,\ \mu_{222}=0.34 $. Now the confidence degree of each IDS group to each of the quantized level of $ y $ should be computed using a T-norm operator; here the minimum operator is chosen for T-norm; So it will be as follows:\\
$ \mu_{11}=min(\mu_{111},\mu_{112})=min(0.17,0.34)=0.17,\ \mu_{12}=min(\mu_{121},\mu_{122})=min(0.34,0.67)=0.34,\ \mu_{21}=min(\mu_{211},\mu_{212})=min(0.67,0.67)=0.67 \ and \ \mu_{22}=min(\mu_{221},\mu_{222})=min(0.34,0.34)=0.34 $.\\
Now, the confidence degree of the system to each of the output level should be computed using a S-norm operator, here the maximum function is used for the S-norm operator; so the system fuzzy output will be computed as follows:\\
$ \mu(y=1)=\mu_1=max(\mu_{11},\mu_{12})=max(0.17,0.67)=0.67 $\\
$ \mu(y=2)=\mu_2=max(\mu_{21},\mu_{22})=max(0.34,0.34)=0.34 $\\
the output fuzzy number of the model will be the pairs $ (y_1,\mu_1)=(1,0.67) $ and $ (y_2,\mu_2)=(2,0.34) $. The crisp output, $ y $, can be computed using the WSF.\\
$ y=\frac{1\times0.67+2\times0.34}{0.67+0.34}=1.35 $\\
So the model estimated output for the input $ (x_1,x_2)=(2.5,3.5) $ is $ y=1.35 $ which is near the quantized level $ y=1 $.\\

\section{A Brief Introduction to memristive Devices }
Three fundamental elements in electrical circuits known as resistor, capacitor and inductor are defined by the relations between two of the four fundamental circuit variables i.e. current, voltage, charge and flux. In 1971, Leon Chua introduced the forth basic element of circuit named Memristor \cite{chua}. After 37 years, in May 2008, Dmitri Strukov et al. at HP Labs published a paper announcing a model for the first physical realization of the Memristor \cite{strukov}. Since then,  many researches have been conducted focusing on applications of Memristor such as non-volatile RAM \cite{memram}, implementation of spiking neural network and emulation of human learning \cite{memlea1,memlea2}, implementation of fuzzy and neuro fuzzy systems \cite{farnood1,farnood2}, building programmable analog circuits \cite{meman1,meman2}, and implementing digital circuits \cite{memdig}.

Memristor is a passive device that provides a functional relation between charge and flux. It is defined as a two-terminal circuit element in which the flux between the two terminals is a function of the amount of electric charge that has passed through the device as defined in Eq. (\ref{5}).
\begin{equation}
\label{5}
d\phi=R_MdQ 
\end{equation}
Differentiating both sides of Eq. (\ref{5}) with respect to time (t), the Eq. (\ref{6}) can be obtained.
\begin{equation}
\label{6}
R_M=\frac{d\phi/dt}{dQ/dt}=\frac{v(t)}{i(t)} 
\end{equation}
Memristor actually behaves like a variable resistor; its resistance can be changed by applying voltage to or passing current through its terminals.
\begin{figure}
\centering
\includegraphics[scale=.18]{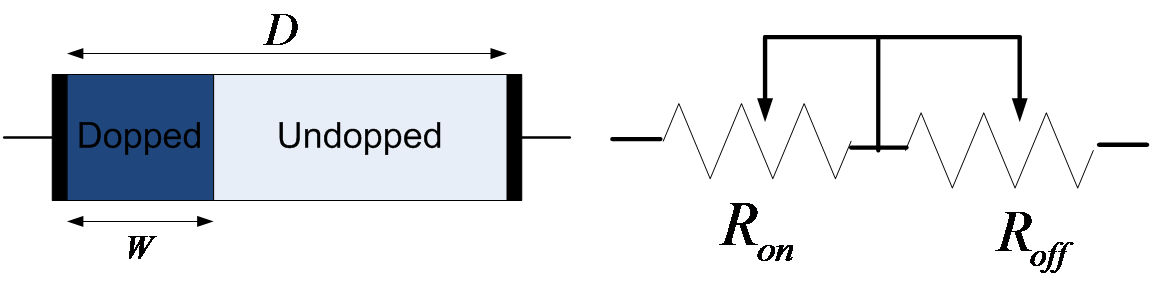}
\caption{$ TiO_2 $ Memristor fabricated in HP lab by Strukov and it's equivalent circuit}
\label{memristor}
\end{figure}

Strukov \textit{et al.} in HP lab realized Memristor by using a very thin film of Titanium Dioxide ($ TiO2 $). As shown in figure \ref{memristor}, the thin film is sandwiched between two platinum ($ pt $) contacts and one side of $ TiO_2 $ is doped with oxygen vacancies. In figure \ref{memristor}, $ D $ is total length of the device and $ w $ determines the length of doped region. The oxygen vacancies are positively charged ions. Thus, there is a $ TiO_2 $ junction where one side is doped and the other side is undoped. Pure $ TiO_2 $ is a semiconductor and has high resistivity. The doped oxygen vacancies make the $ TiO_{2-x} $ material much more conductive compared to $ TiO_2 $. If some electric charge passes through the device, $ w $ will change. If $ w=D $, the device will have its minimum resistance hereafter called $ R_{on} $ and when $ w=0 $, it has its maximum resistance denoted by $ R_{off} $. 

The mathematical model of HP memristor is as Eq. (\ref{7}) \cite{strukov}.
\begin{eqnarray}
\label{7}
w(t)=w_0+\frac{\mu_vR_{on}}{D}q(t) ,  \nonumber \\
R_M(w)=R_{on}\frac{W}{D}+R_{off}(1-\frac{W}{D})
\end{eqnarray}
Where $ w_0 $ is the initial width of the doped region, $ \mu_v $ is the average ion mobility and $ q(t) $ is the amount of electric charge (integral of current) that has passed through the device. It is obvious from these equations that passing current in one direction for longer period of time will change the memristance of the memristor more. Moreover, by setting the passing current to zero, the memristance will not change anymore implying that the device can act as a memory. It should be noted that the direction of the current is an important factor. Passing electric charge in one direction will reduce the resistance, while changing the direction of current will increase the resistance of the device. So the amplitude, the direction and the duration of the passing current are the parameters which affect the amount of change in the memristance. Note that determining the memristance at any time can be done by applying a small voltage below a threshold ($ V_{th} $) through the memristor and measuring the passing current through it.

\section{Hardware implementation of the proposed algorithm based on memristor-crossbar structure}
Hardware implementation of the proposed algorithm consists of three parts: control unit, IDS plane and inference algorithm implementation. The control unit which controls the timing and sequence of the procedure can be implemented using any programmable hardware like microcontrollers, FPGA or DSP processors. In this work the implementation of control unit will not be discussed in details and just its task will be defined.

\begin{figure*}
\centering
\includegraphics[scale=.43]{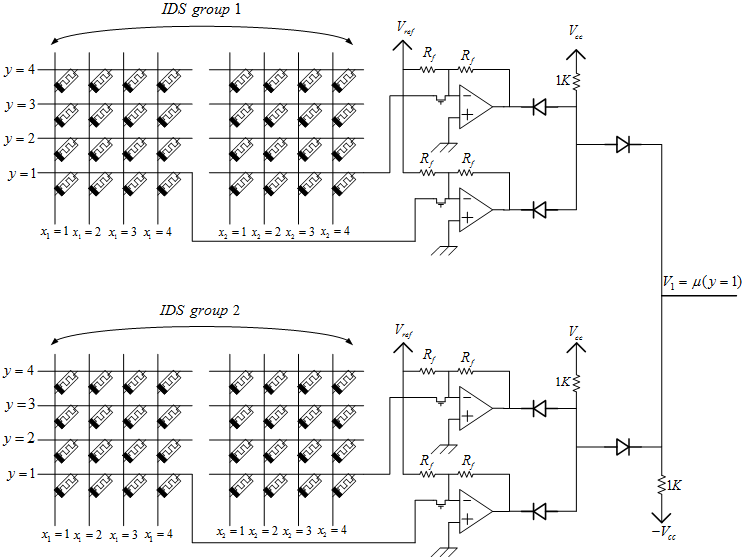}
\caption{Proposed hardware for a system with two inputs and one output}
\label{memristorcrossbar}
\end{figure*}

For implementation of the IDS planes the memristor-crossbar structure which is proposed in \cite{farnood1} is used. Each IDS plane with resolution $ n_r\times n_c $ can be shown by a memristor-crossbar structure with $ n_r $ rows and $ n_c $ columns. In this case, each memristor corresponds to a pixel in the IDS plane and hence its value can be considered as the pixels value. In figure \ref{memristorcrossbar} the IDS planes and the inference part for one quantized level of output variable is shown. It is assumed that the system is composed of two IDS groups, each with two IDS planes.
In the learning phase, the N-MOS transistors are off so the inference circuit will be separated from the IDS planes. For programming the memristor-crossbar planes, a learning pulse should be applied to the column corresponding to the input variable quantized level and the row correspondence to the quantized level of the output value should be set to zero voltage. In \cite{farnood1}, authors have shown that this procedure will lead to a semi-Laplacian form of the ink stains on the memristor-crossbar structures. But the control unit should select a new plane group for each new training data; this can be done by some switches and multiplexers easily. The initial memristance of the memristors in the crossbar structures is  set to the minimum value named $ R_{on} $. So by applying the learning pulse the memristance will be increased. It should be noted that the amount of the changing in the memristance of the memristors is related to amplitude and width of the learning pulse. Wider pulses and bigger amplitudes will change the memristance more.

\begin{figure}
\centering
\includegraphics[scale=.18]{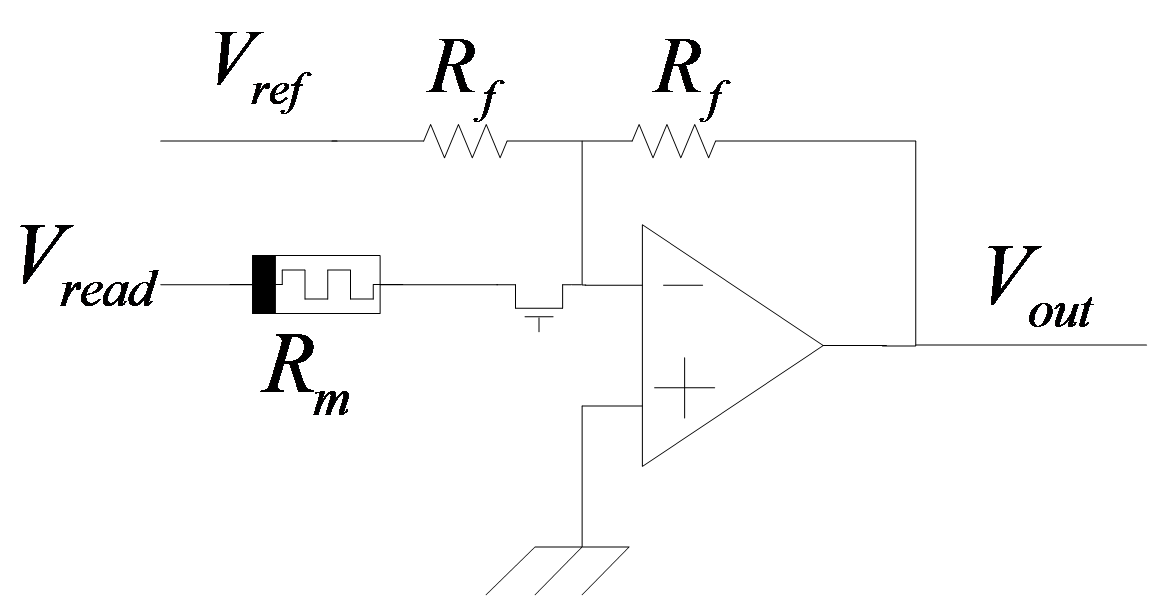}
\caption{Inverting amplifier structure which is used for translating the memristance to the voltage variable.}
\label{voltagetranslatecircuit}
\end{figure}

In order to  implement  the inference part, in the first step the memristance of the memristors should be translated to current or voltage variable. In this paper we used the voltage variable. In figure \ref{voltagetranslatecircuit} the Op-Amp is used as an adder with inverted output so its output can be computed from Eq. (\ref{8}).
\begin{equation}
\label{8}
v_{out}=-v_{ref}+v_{read}(-\frac{R_f}{R_m})
\end{equation}

Where $ v_{read} $ is a voltage which is applied to the column of the IDS plane related to quantized level of the input value; its amplitude is set to less than $ v_{th} $ of the memristors to hold the memristance of the memristors unchanged during the modeling phase, $ v_{ref} $ is a reference voltage with negative polarity, $ R_f $ is the feedback resistor with a value equal to $ R_{on} $ , and finally $ R_m $  is the memristance of the memristor. The output voltage of the Op-Amp,  $ v_{out} $ , is the confidence degree of the relative IDS plane to the corresponding value of output. As can be inferred from the equation …  $ v_{ref} $ should be chosen equal to $ v_{read} $  with negative polarity, this leads to the result that when the memristors are in their initial state the output voltage of op-amp will be zero as shown in Eq. (\ref{9}).
\begin{eqnarray}
\label{9}
&v_{out}&=-v_{ref}+v_{read}(-\frac{R_{on}}{R_{on}})=-v_{ref}-v_{read} \nonumber \\
&=&-(-v_{read})-v_{read}=0  
\end{eqnarray}
This means that the confidence degree of the corresponding IDS plane to this value of $ y $ is zero or in other words this IDS planes has no information about the corresponding value of $ y $. In the learning procedure the memristance of the memristors may increase in which case the absolute value of the $ v_{read}(-\frac{R_f}{R_m}) $  will strat decreasing and as a result $ -v_{ref}+v_{read}(-\frac{R_f}{R_m}) $ will become larger. So when the output voltage of one Op-Amp becomes larger, it means that the confidence degree of the corresponding IDS plane to the relative $ y $ value is higher.

\begin{figure}
\centering
\includegraphics[scale=.5]{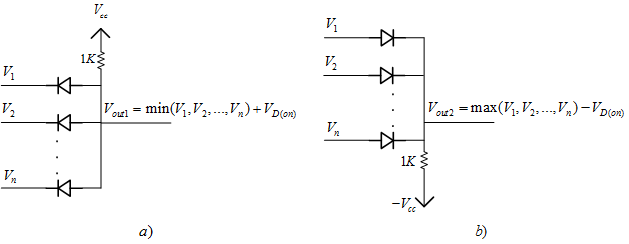}
\caption{a) Implementation of the T-norm operator with a min circuit; b) Implementation of the S-norm operator with a max circuit.}
\label{minmax}
\end{figure}

For implementation of the minimum and maximum functions, some very simple circuits using diodes are used. In \ref{minmax} the schematics of both circuits are shown. The figure \ref{minmax} (a) shows the implementation of the minimum function as T-norm operator. The output voltage of this circuit is $ min(V_1,V_2)+V_{D(on)} $; So the output voltage will be the minimum voltage of the inputs plus a bias voltage equal to $ V_{D(on)} $ ( $ V_{D(on)} $ is the forward bias voltage of diode). In figure \ref{minmax} (b)  the maximum circuit is shown. The output voltage of this circuit is $ max(V_1,V_2)-V_{D(on)} $ ; so the output of this stage will be the maximum voltage of its inputs minus a bias voltage equal to $ V_{D(on)} $ . Because the outputs of the minimum stages are the inputs of the maximum stages, the output of the circuit will have no bias.
The output of the maximum stage will be the confidence degree of the system to the corresponding value of $ y $. so if the quantization resolution of the variable $ y $ is $ n_y $,there will be $ n_y $ voltages, and each is confidence degree of the system to its corresponding value of $ y $. 

\begin{figure}
\centering
\includegraphics[scale=.4]{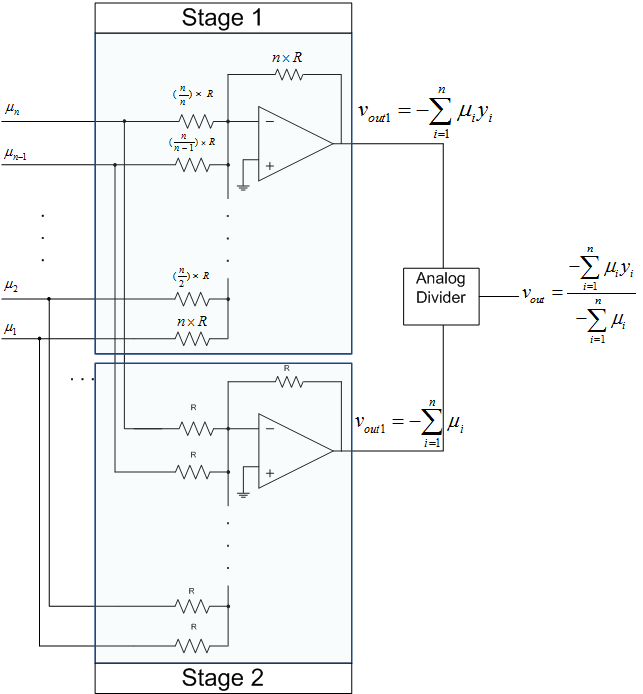}
\caption{Proposed circuit for implementation of the WAF as the defuzzification stage.}
\label{Defuzzification1}
\end{figure}

For implementing the WAF algorithm as defuzzification stage, the nominator and denominator of the WAF are calculated separately and finally divided using an analog voltage divider circuit. This part of the hardware consists of two stages as shown in figure \ref{Defuzzification1} . The first stage consists of an inverting amplifier structure in which the feedback resistor is set to $ n_y \times R $   where $ n_y $ is the resolution of the quantization and $ R $ is a resistor whose value can be between 1 to 10 Kilo Ohm depending on the technology of the fabrication. The voltage $ v_i=\mu_i $  is the output voltage of the \textit{i}th S-norm cell. The output voltage of stage 1 will thus be equal to $ -\sum \limits_{i=1}^n (\mu_i \times y_i) $ as shown in Eq. (\ref{vout}).\\
$ v_{out_1}= $
\begin{eqnarray}
\label{vout}
-(\mu_1 \frac{(n \times R)}{(\frac{n}{1} \times R)}+ \mu_2 \frac{(n \times R)}{(\frac{n}{2} \times R)}+ ...+\mu_n \frac{(n \times R)}{(\frac{n}{n} \times R)})\nonumber \\ 
=-(\mu_1 \times 1+\mu_2 \times 2+...+\mu_n \times n) \nonumber \\
=-\sum \limits_{i=1}^n (\mu_i \times i)=-\sum(\mu_i \times y_i)\nonumber\\
\end{eqnarray}
Similarly the output voltage of the second stage can be calculated using Eq. (\ref{vout2}).
\begin{equation}
\label{vout2}
v_{out2}=-(\mu_1+\mu_2+...+\mu_n)=-\sum \limits_{i=1}^n \mu_i
\end{equation}
Applying these voltages to the inputs of an analog voltage divider, the output voltage will be equal to $ v_{out}=\frac{-\sum\limits_{i=1}^n \mu_i y_i}{-\sum\limits_{i=1}^n\mu_i} = \frac{\sum \limits_{i=1}^n \mu_i y_i}{\sum \limits_{i=1}^n \mu_i}$ , which is the WAF of the output fuzzy number.

\section{Simulation results}
In order to investigate the efficacy of the proposed algorithm, we tested it in two different applications. The first application is the modeling of two complex functions which are used as a benchmark in many relative studies. The next application is the classification task performed on three different sets of data. All simulations are conducted in MATLAB version 2012 software.
For the modeling test, two functions $ F_1 $ and $ F_2 $ which are defined in Eq.(\ref{testfunction}) are selected.
\begin{eqnarray}
\label{testfunction}
F_1=(1+x_1^{-2}+x_2^{-1.5})^2 ,\ 1<x_1,x_2<10,\nonumber \\
F_2=\sqrt{2(\frac{sin(x_1)}{x_1})^2+3(\frac{sin(x_2)}{x_2})^2} ,\ 1<x_1,x_2<10,\nonumber \\
\end{eqnarray}

The Fraction of Variance Unexplained (FVU) index which is defined in Eq. (\ref{fvu}) is used for evaluation of the performance of the proposed algorithm.
\begin{equation}
\label{fvu}
FVU=\frac{\sum\limits_{l=1}^L(\hat{y}(x_l))-y(x_l))^2}{\sum\limits_{l=1}^L(y(x_l)-\bar{y}(x_l))^2}
\end{equation} 
Where $ \hat{y} $ is the output of the constructed model, $ \bar{y}=\frac{1}{L}\sum\limits_{l=1}^L y(x_l) $ and $ L $ is number of the sample data. The FVU is proportional to the mean squared error; as the model's accuracy increases, the FVU approaches zero. 
In Table \ref{mytable} the results of simulations are shown. The simulations are done with 250, 550 and 1000 training data with different radiuses of the ink stains 10, 20 and 30, where this radius of ink stains is the same for the inputs and output variables. Each variable is quantized to 128 levels in all cases.

\begin{table}
\caption{FVU index for modeling task}
\centering
\label{mytable}
\begin{tabular}{c c c c c}
\hline
FVU \ index&&250&550&1000 \\ \hline
&$ F_1 $&$ NAN $&0.0780&0.0740\\
R=10&&&&\\
&$ F_2 $&$ NAN $&0.0315&0.0167\\ \hline
&$ F_1 $&0.2120&0.1953&0.2064\\
R=20&&&&\\
&$ F_2 $&0.0862&0.0638&0.0487\\ \hline
&$ F_1 $&0.3270&0.3642&0.3539\\
R=30&&&&\\
&$ F_2 $&0.1831&0.1507&0.1193 \\ \hline
\end{tabular} 
\end{table}

As can be seen from the results, FVU index is very low and the proposed algorithm has the generalization ability. By decreasing the radius of the inks stains the accuracy of the model increases provided that we have enough number of training data; otherwise, the radius of inks should be increased. It is worth to mention that increasing the number of training data to infinity and decreasing the radius of the ink's stains to zero will change the algorithm to a rich look-up table.

\begin{figure}
\centering
\includegraphics[scale=.35]{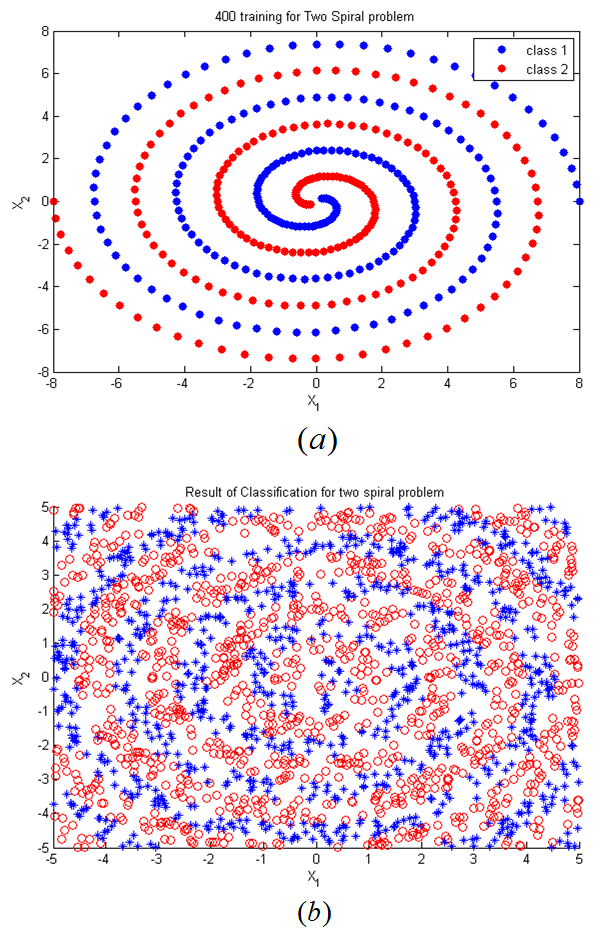}
\caption{a) Training data for two spiral problem; b) result of classification using proposed algorithm.}
\label{twospiral}
\end{figure}

For evaluating the algorithm in classification task three problems are chosen; the first one is two-spiral problem. The two-spiral problem is a well-known classification benchmark for supervised learning algorithms \cite{22}. For solving this problem, we assumed it as a two input-one output function in which the value of the output variable for the points on one spiral is 1 and for the other is 2; 400 training data was used for training the model, resolution of quantization for the input and output variables was set to 128 and 2 respectively. In figure \ref{twospiral} the result of classification using proposed algorithm is shown. The blue dots show the points which belong to class 1 and the red dots show the points which are classified in class 2. The classification is done with 100 percent accuracy.

\begin{figure}
\centering
\includegraphics[scale=.35]{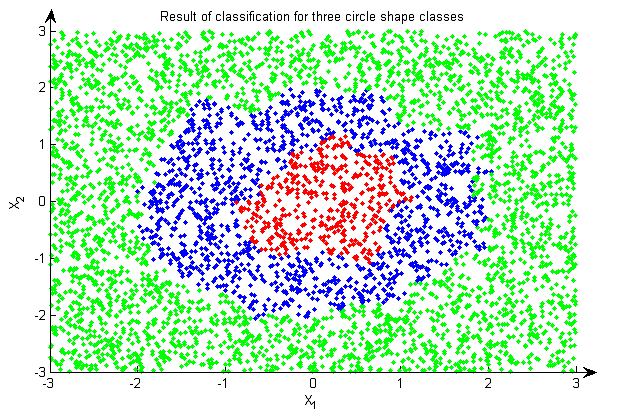}
\caption{Result of classification for three circle shape classes.}
\label{circle}
\end{figure}

In the next case three classes are defined which form the Eq. (\ref{classcircle}). In this case 300 training data is used for construction of the model. Resolution of the output variable is set to 32 , resolution of the input variables was set to 256, and the radius of the inks stains for the input variables  is set to 50 and for the output variable is set to 16. The result of the classification is shown in figure \ref{circle} . The average of the classification accuracy is 98.7 percent for 100 times running the task.

\begin{eqnarray}
\label{classcircle}
&class1:& \ x_1^2+x_2^2<1  \nonumber \\
&class2:& \ 1<x_1^2+x_2^2<4 \nonumber \\
&class3:& \ 4<x_1^2+x_2^2
\end{eqnarray}

In the last case the iris dataset is used as the benchmark for the evaluation of the proposed algorithm \cite{23}. The dataset is composed of 150 data samples of three types of the iris flower and each one is defined by four features. There are 50 samples of each type in the dataset. For investigating the accuracy of the algorithm, the result is compared with the Multi-Layer Perceptron (MLP) and ANFIS algorithms. The MLP structure has two layers with 10 neurons in its hidden layer. The simulation is done for 100 times for each algorithm and the average of the results is reported. For the simulation, 100 sample data are chosen randomly as the training data and 50 sample data was used for test data. As can be seen in Table \ref{mytable1} our algorithm works better than both MLP and ANFIS in the classification of the Iris data. It should be noticed that the result of the proposed algorithm is sensitive to the radius of the inks stains and the resolution of the quantization for the input and output variables, setting this parameters in a proper range needs some trial and error which is a disadvantage of our algorithm with respect to MLP and ANFIS algorithms.

\begin{table}
\caption{Comparison the performance of proposed algorithm with MLP and ANFIS in classification task}
\centering
\label{mytable1}
\begin{tabular}{c c c c}
\hline
&proposed \ algorithm&MLP&ANFIS \\ \hline
Performance(percent)& 96.05 & 92.79& 94.02 \\
\hline
\end{tabular} 
\end{table}

\section{Discussion and conclusion}
In this paper we proposed a new modeling algorithm based on ALM concept. The structure of the system is discussed first, then the \textit{if then} rules for the rule base of the fuzzy inference system has been written, and the inference algorithm of the proposed structure has been developed. The algorithm has no need of optimization and is not dependent on the order of the learning sample data. A simple hardware based on memristor-crossbar structure is proposed for the hardware implementation of the algorithm. The simulation results have shown that our algorithm is very effective in modeling and classification tasks.

\begin{figure}
\centering
\includegraphics[scale=.55]{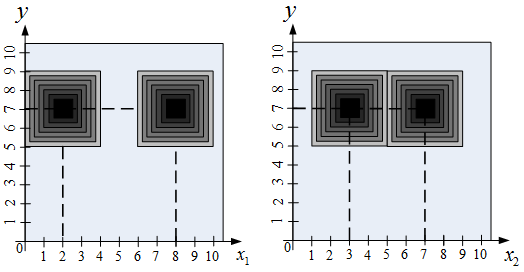}
\caption{Pictorial demonstration for invalid situation.}
\label{notallowedsituation}
\end{figure}

Although the efficacy of the algorithm is very high, the memory which is used for implementation of the IDS planes is so huge and is proportional with the number of input variable and number of training data. There is no way for reducing the number of input variables in this algorithm, so for reducing the number of the IDS groups a simple rule is developed, considering an IDS group for each input data which the models output is not equal to the desired output. This rule will decrease the number of the IDS groups effectively. Another suggestion for reducing the number of IDS groups is saving multiple training data on one IDS groups. for aggregation of the neighboring inks stains any S-norm operator can be chosen. There is a rule for diffusing multiple training on the same IDS group which says never save two sample data whose output variables are equal. In figure \ref{notallowedsituation}, the situation is shown. Assume two data samples $ (x_1,x_2,y)=(2,7,7) $ and $ (x_1,x_2,y)=(8,3,7) $ are diffused on same IDS group. Although there are no training data near the points like $ (x_1,x_2)=(2,3) $ or $ (x_1, x_2)=(7,8) $ , this IDS group will result in the maximum confidence degree for the output value $ y=7 $ for both of these input points which are  clearly wrong.

As it can be seen in the simulation results, the output is very dependent on the radius of the ink stains on the IDS planes. If the radius of inks becomes bigger, the accuracy will be less and vice versa. But there is a relationship with the density of the training data in the space. If the density is high, the radius can be smaller and when the density of sample data is low in any of points of the space, the radius should be more. If the radius isn't big enough, the confidence degree will be zero for all quantized values of the output variable for the input points which are not in the range of the $ R $ neighbor of the training data. As a result, it can be deduced that the radius can be adaptively chosen with respect to the density of the data in any part of the space. Developing an adaptive learning algorithm for this structure will be our future work. 

\bibliographystyle{model1a-num-names}

\begin{thebibliography}{0}

  
\bibitem{zadeh1} L.A. Zadeh, Outline of a new approach to the analysis of complex systems and decision processes, \textit{IEEE Trans. Systems, Man, and Cybernetics} \textbf{1} (1973), 28--44.

\bibitem{zadeh2} L.A. Zadeh, Soft computing and fuzzy logic, \textit{Software, IEEE} \textbf{Vol. 11} (1994), 48--56.


\bibitem{sugeno} T. Takagi and M. Sugeno, Fuzzy identification of systems and its application to modeling and control, \textit{IEEE Trans. Systems, Man and Cybernetics(1)} \textbf{15} (1985), 116--132.

\bibitem{tanaka} K. Tanaka, T. Ikeda and H. O. Wang, Design of fuzzy control systems based on relaxed LMI stability conditions, \textit{Proceedings of 35th IEEE Conference on Decision and Control} \textbf{Vol. 1} (1996), 598--603.


\bibitem{shibata} S. Kondo, T. Shibata, and T. Ohmi, Superior generalization capability of hardware-learning algorithm developed for self-learning neuron MOS neural networks, \textit{Jpn. J. Appl. Phys} \textbf{Vol. 34} (1995), 1066-1069.

\bibitem{degaris} H. de Garis, CAM-Brain : Growing an Artificial Brain with a Million Neural Net Modules inside a Trillion Cell Cellular Automata Machine, \textit{Journal of the Society of Instrument and Control Engineers (SICE)} \textbf{Vol.33, No.2} (1994).

\bibitem{strukov} D.B. Strukov, G.S. Snider, D.R. Stewart and R.S. Williams, The missing memristor found, \textit{Nature} \textbf{453} (May 2008), 80--83.

\bibitem{bernabe1} B. Linares-Barranco and T. Serrano-Gotarredona, Memristance can explain Spike-Time-Dependent-Plasticity in Neural Synapses, \textit{available from Nature Precedings}, \textbf{} {http://hdl.handle.net/10101/npre.2009.3010.1, March, 2009}.

\bibitem{bernabe2} C. Zamarreño-Ramos, L. A. Camuñas-Mesa, Jose A. Perez-Carrasco, T. Masquelier, T. Serrano-Gotarredona, and B. Linares-Barranco, On Spike-Timing-Dependent-Plasticity, Memristive Devices, and building a Self-Learning Visual Cortex, \textit{Frontiers in Neuromorphic Engineering} \textbf{} (2011), 5-26.

\bibitem{farnood1} F. Merrikh-bayat, S. B. Shouraki and  A. Rohani, Memristor Crossbar-based Hardware Implementation of IDS Method, \textit{IEEE Transaction on Fuzzy Systems} \textbf{19} (Dec. 2011), 1083--1096.

\bibitem{alm1} S.B. Shouraki and N. Honda, Recursive Fuzzy Modeling Based on Fuzzy Interpolation, \textit{Journal of Advanced Computational Intelligence} \textbf{3} (April 1999), 114--125.

\bibitem{farnood2} F. Merrikh-Bayat, S. Bagheri Shouraki, Memristive Neuro-fuzzy System, \textit{IEEE Transactions on Systems, Man and Cybernetics, Part B} \textbf{Vol. 43} (Feb. 2013), 269--285.

\bibitem{almmod} M. Murakami and N. Honda, A study on the modeling ability of the IDS method: A soft computing technique using pattern-based information processing, \textit{International Journal of Approximate Reasoning} \textbf{45} (2007), 470--487.

\bibitem{almclas} M. Murakami and N. Honda, Classification performance of the IDS method based on the two-spiral benchmark, \textit{Proceedings of the 2005 IEEE International Conference on Systems, Man and Cybernetics (SMC'05)} \textbf{} (Oct. 2005), 3797--3803.


\bibitem{almcon1} S. A. Shahdi and S. B. shouraki, Supervised Active Learning Method as an Intelligent Linguistic Controller And Its Hardware Implementation \textit{in Proc. the Second IASTEAD International Conference on Artificial Intelligence And Applications} \textbf{} (Spain, 2002).

\bibitem{almcon2} Y. Sakurai, \textit{A Study of the Learning Control Method Using PBALM-a Nonlinear Modeling Method}, Ph.D. Dissertation, 2005.

\bibitem{almop1} M. Murakami and N. Honda, A basic constructive algorithm for the IDS method, \textit{Proceedings of the Joint 3rd International Conference on Soft Computing and Intelligent Systems and 7th International Symposium on Advanced Intelligent Systems} \textbf{} (September 2006), 355--360.

\bibitem{almop2} H. Sagha, S. Bagheri Shouraki, H. Beigy, H. Khasteh and E. Enayati, Genetic Ink Drop Spread, \textit{Second International Symposium on Intelligent Information Technology Application} \textbf{} (2008).



\bibitem{chua} L.O. Chua, Memristor - the missing circuit element, \textit{IEEE Trans. on
Circuit Theory} \textbf{18} (1971), 507--519.



\bibitem{memram} R. Waser, and M. Aono, Nanoionics-based resitive switching memories, \textit{Nature Materials} \textbf{6} (2007), 833--840.

\bibitem{memlea1} Y.V. Pershin, S.L. Fontaine and M.D. Ventra, Memristive model of amoeba's learning, \textit{Phys. Rev. E} \textbf{80} (2009).

\bibitem{memlea2} J. A. Perez-Carrasco, C. Zamarreo-Ramos, T. Serrano-Gotarredona, and B. Linares-Barranco, On neuromorphic spiking architectures for asynchronous STDP memristive systems, \textit{Proceedings of IEEE international symposium on circuits and systems} \textbf{} (2010), 1659--1662.


\bibitem{meman1} Y. V. pershin, and M.D. Ventra, Practical Approach to Programmable Analog Circuits With Memristors, \textit{IEEE Transactions on Circuits and Systems I: Regular Paper} \textbf{57} (Aug. 2010), 1857--1864.

\bibitem{meman2} S. Shin, K. Kim, and S.M. Kang, Memristor-based fine resolution resistance and its applications, \textit{International Conference on Communications, Circuits and Systems, 2009. ICCCAS 2009} \textbf{ } (July 2009), 948--951.

\bibitem{memdig} P. Kuekes, Material Implication: digital logic with memristors, \textit{Memristor and Memristive Systems Symposium} \textbf{ } (Nov. 2008).

\bibitem{22} K. J. Lang and M. J.Witbrock, Learning to tell two spirals apart, \textit{Proc. 1988 Connectionist Models Summer School} \textbf{ } (1988), 52--59.

\bibitem{23} \textit{http://archive.ics.uci.edu/ml/datasets/Iris}.



\end{thebibliography}

\end{document}